\documentclass[journal]{IEEEtai}

\usepackage[colorlinks,urlcolor=blue,linkcolor=blue,citecolor=blue]{hyperref}

\usepackage[numbers,sort&compress]{natbib}

\usepackage{color,array}
\usepackage{booktabs}
\usepackage{graphicx}
\usepackage{amsmath}
\usepackage{multirow}
\usepackage{pifont}

\usepackage{lineno}

\begin{document}

\title{Decoupling Dark Knowledge via Block-wise Logit Distillation for Feature-level Alignment} 

\author{%
  Chengting Yu$^{1,2}$, Fengzhao Zhang$^{2}$, Ruizhe Chen$^{2}$, Aili Wang$^{1,2,*}$, Zuozhu Liu$^{2}$, Shurun Tan$^{1,2}$, Er-Ping Li$^{1,2}$  \\
  $^1$ College of Information Science and Electronic Engineering, Zhejiang University\\
  $^2$ ZJU-UIUC Institute, Zhejiang University\\
  chengting.21@intl.zju.edu.cn, ailiwang@intl.zju.edu.cn
}

\maketitle

\begin{abstract}
Knowledge Distillation (KD), a learning manner with a larger teacher network guiding a smaller student network, transfers dark knowledge from the teacher to the student via logits or intermediate features, with the aim of producing a well-performed lightweight model. 
Notably, many subsequent feature-based KD methods outperformed the earliest logit-based KD method and iteratively generated numerous state-of-the-art distillation methods.
Nevertheless, recent work has uncovered the potential of the logit-based method, bringing the simple KD form based on logits back into the limelight.
Features or logits? They partially implement the KD with entirely distinct perspectives; therefore, choosing between logits and features is not straightforward.
This paper provides a unified perspective of feature alignment in order to obtain a better comprehension of their fundamental distinction. 
Inheriting the design philosophy and insights of feature-based and logit-based methods, we introduce a block-wise logit distillation framework to apply implicit logit-based feature alignment by gradually replacing teacher's blocks as intermediate stepping-stone models to bridge the gap between the student and the teacher.
Our method obtains comparable or superior results to state-of-the-art distillation methods.
This paper demonstrates the great potential of combining logit and features, and we hope it will inspire future research to revisit KD from a higher vantage point.
\end{abstract}



\section{Introduction}
\label{sec:intro}
\IEEEPARstart{D}{eep} neural networks have shown exceptional performance in several domains, including computer vision \cite{huang_densely_2018,he_deep_2016,he2017mask}
and natural language processing 
\cite{devlin_bert_2019, otter2020survey}.
Nevertheless, it is common for robust and powerful networks to get advantages from large model capacity, hence resulting in significant computational and storage expenses. 
Various strategies have been developed to train fast and compact neural networks that are both efficient in terms of speed and size, including the invention of novel architectures \cite{cui_fast_2019, howard_searching_2019, sandler_mobilenetv2_2018,zhang_shufflenet_2018}, network pruning
\cite{li_pruning_2017,frankle_lottery_2019,liu_rethinking_2019}, quantization \cite{jacob_quantization_2018}, and knowledge distillation \cite{hinton_distilling_2015, romero_fitnets_2015, wang_knowledge_2022}.
Among the investigated strategies, knowledge distillation (KD) has emerged as one of the most promising techniques to obtain a lightweight model with practicality and efficiency. KD refers to a set of methods concentrating on transferring “dark knowledge” from a large model (teacher) to a light one (student), which are applicable to a variety of network architectures and can be combined with numerous other techniques, such as network pruning and quantization \cite{wei_quantization_2018} toward lightweight models.

\begin{figure}[t]
\centering
\includegraphics[width=9cm]{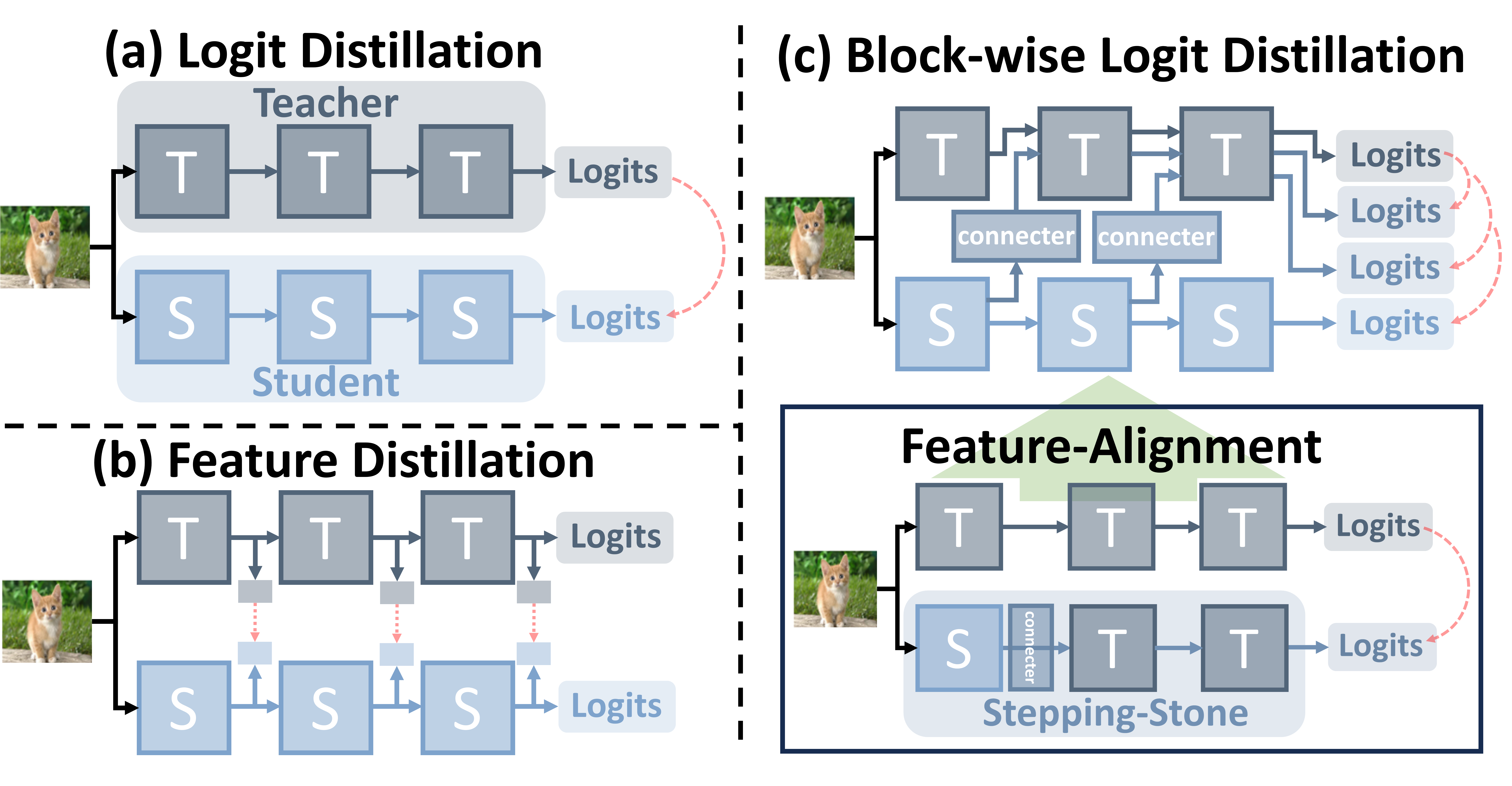} 
\caption{Method Illustration. (a) Logit distillation transfers the entire dark knowledge based solely on output logits. (b) Feature distillation further implements feature alignment at multiple levels. (c) The proposed block-wise logit distillation framework accomplishes implicit feature alignment via output logits of stepping-stones. We refine the distillation into logit-only objectives, with the transfer of dark knowledge decoupled at the block level.}
\label{fig1}
\end{figure}

The primary approaches used in KD can be categorized into two main streams: 1) logit distillation and 2) feature distillation. The earliest concept of KD was proposed based on logits \cite{hinton_distilling_2015}, which transfers only logit-level knowledge from teacher to student by minimizing the KL-Divergence between prediction logits (Fig. \ref{fig1}a).  
Since then, towards better utilization of teacher knowledge \cite{romero_fitnets_2015}, the majority of KD study has been dedicated to investigating intermediate layers and performing distillation via the alignment of feature distributions \cite{heo_comprehensive_2019,chen_distilling_2021}, which are referred to as feature distillation (Fig. \ref{fig1}b). 
Nevertheless, recent work \cite{zhao_decoupled_2022} reveals and improves the limitations of the classical KD loss and re-examines the potential of the logits, bringing the simple logit-based KD back into the limelight.

Logits or features? Although both logit-based and feature-based methods implement the distillation, they do so with quite different perspectives on the "dark knowledge", deciding between the two very hard. It's time to gaze at the gap between them to discover an avenue to connect the two lines together.
For logit-based methods, "dark knowledge" is thought to be fully contained in the output logits, so aligning the logits is sufficient to transfer the "dark knowledge" from teacher to student (Fig. \ref{fig1}a); however, for features-based methods, this alignment requirement is extended to the hidden features across different block-levels in an effort to transfer different levels of "dark knowledge" by aligning block-wise features (Fig. \ref{fig1}b).
Clearly, the features-based methods are more hypothetical and implemented with more stringent constraints in the hopes of facilitating the transfer of "dark knowledge" to the student at finer levels; as a result, this approach may frequently converge faster and achieve a good performance.
Meanwhile, in some cases, end-to-end logit-based alignment is more secure for transferring "clear" knowledge since features-based alignment may also preserve task-irrelevant noise items throughout the distillation process. 
We argue that, although feature-based approaches have more success than logit-based methods at conveying "dark knowledge" through block-wise alignments, they are restricted by the need for excessively stringent alignment constraints.

Inspired by this observation, we propose a new distillation framework called block-wise logit distillation, which implements the alignment of feature-level “dark knowledge” based on the unified logit-based distillation of intermediate stepping-stone models (Fig. \ref{fig1}c) to fill in the gap between features and logits. As the implicit intermediate models during training, the student's shallow blocks are gradually substituted with the teacher's to establish stepping-stones, which are eventually abandoned during inference. We further evaluate the proposed methods on several visual benchmark datasets, including CIFAR100, ImageNet, and MSCOCO datasets for image classification and object detection tasks. The experimental results demonstrate that our method consistently outperforms both logit- and feature-based distillation methods. 
{Further experiments on natural language processing tasks based on the BERT model have also been included in the experimental section to demonstrate the versatility of our approach.}
{
Additionally, a comparative analysis of training costs against related distillation methods yields several key insights:
1. Compared to feature-based KD methods, our approach not only achieves superior performance but does so with a similar degree of added complexity, highlighting the efficiency of the proposed framework without significantly increasing computational demands.
2. Regarding adaptability, we have formulated a lightweight version of the original framework that considerably diminishes the complexity typical of the standard method while still preserving high accuracy, which offers a feasible solution for contexts where computational resources are scarce.
3. The framework is compatible with recent logit-based KD approaches, providing an alternative option that can further enhance performance, which underscores the versatility of our method, rendering it an invaluable asset for augmenting the effectiveness of logit-based knowledge distillation.
}

\section{Related Work}
\label{sec:related}

The concept of knowledge distillation \cite{hinton_distilling_2015} was first proposed as a learning strategy that employs a larger teacher network to steer the training process of a smaller student network for various tasks \cite{li_mimicking_2017,li_online_2021}. The working mechanism of vanilla KD is often comprehended via the process of "teaching" \cite{wang_knowledge_2022}, whereby the transfer of "dark knowledge" occurs through the use of soft labels, known as logits, provided by the teacher to the student. Recent studies examine how to select, express, and transfer the "dark knowledge" more practically and effectively, which can be categorized into two main types: logit- and feature-based distillations.

\textbf{Logits Distillation.} In the wake of the earliest KD method based on temperature-regulated distillation \cite{hinton_distilling_2015}, previous logit-based methods have concentrated mainly on effective regularization and optimization methods.
DML \cite{zhang_deep_2018} proposes a mutual learning method to train students and teachers simultaneously. TAKD \cite{mirzadeh_improved_2019} proposes using intermediary "teacher assistants" to transmit knowledge in a step-by-step manner in order to narrow the performance disparity between teachers and students. Additionally, several works \cite{phuong_towards_2019,cheng_explaining_2020} focus on interpreting the principles underlying KD. Recently, DKD \cite{zhao_decoupled_2022} introduces an improved logit-based objective by decoupling the classical KD loss, which re-explores the potential of logit-based methods with comparable performance gains.

\textbf{Feature Distillation.} To further enhance knowledge distillation, feature distillation is proposed to perform alignments on intermediate features as well as logit outputs, which can directly transfer teacher representations \cite{heo_comprehensive_2019, heo_knowledge_2019, romero_fitnets_2015, chen_distilling_2021}
or the 
correlation \cite{park_relational_2019,tian_contrastive_2022,tung_similarity-preserving_2019,peng_correlation_2019}
from the teacher to the student.
Feature methods are more likely to obtain high performance with extensive information from the teacher; however, tight feature alignment frequently relies on prior empirical observation and meticulous adjustment of hyperparameters.

\section{Methedology: Block-wise Logit Distillation}
\label{sec:method}

\begin{figure*}[tb]
\centering
\includegraphics[width=\textwidth]{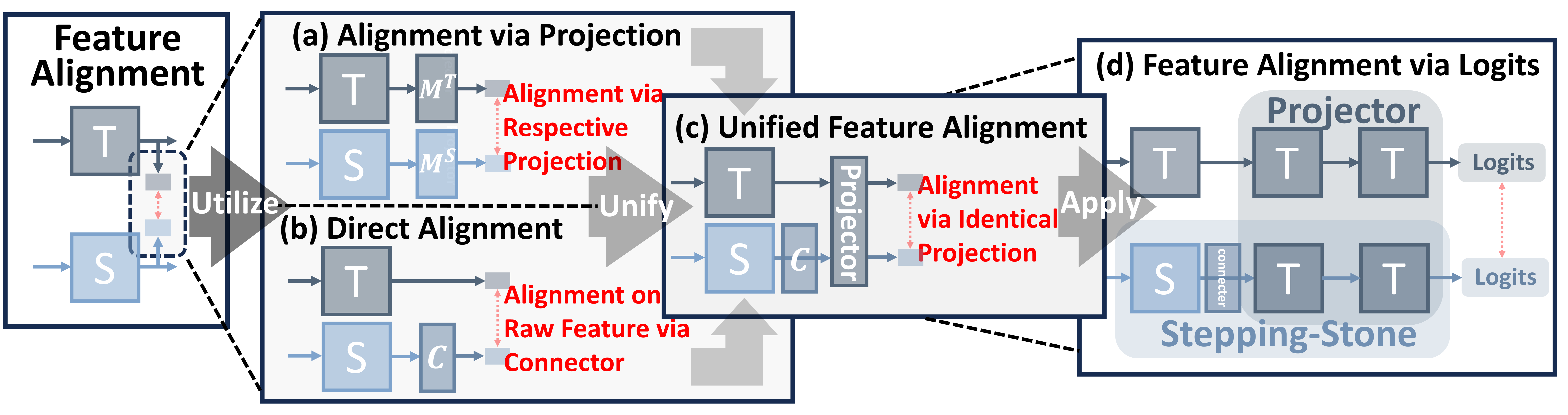} 
\caption{Implementations of Feature Alignment. (a-b) Methods typically employed in feature distillation. (c) The proposed consolidation of alignments with identical projections. (d) Implicit feature alignment utilizing logit distillation with stepping stones. The further extension of implicit logit-based feature alignment is in Fig. \ref{fig3}.}
\label{fig2}
\end{figure*}

\subsection{Background and Notations} 
Given an input image $X$ with targets $Y$, we let $Y^S=S(X)$ and $Y^T=T(X)$ represent the output logits of the student $S$ and the teacher $T$, with
\begin{equation}
    S=\{S_1,S_2,...,S_n,S_c \},T=\{T_1,T_2,...,T_n,T_c\}
\end{equation}
where $n$ denotes the number of blocks, $S_i$ and $T_i$ represent $i$-th block separated by downsampling layers, $S_c$ and $T_c$ represent the fully connected layers that are employed for the final classification.
Then, the process of generating intermediate features $F^S=\{F_i^S\}_{i\leq n}$ can be denoted as:
\begin{equation}
    F_i^S=S_i \circ ... \circ S_1(X)
\end{equation}
We refer to $\circ$ as nesting of functions where $g \circ f(x) = g(f(x))$. 
Given the task objective based on student output $Y^S$ and the ground-truth targets $Y$, we define the cross-entropy loss for the classification task:
\begin{equation}
    L_{task}=CE(softmax(Y^S),Y)
\end{equation}
In knowledge distillation, an extra objective $L_{distill}$ is determined to measure the distance between $S$ and $T$.
As for logit distillation, one tries to match the softened outputs logits of student and teacher via a given distance measurement, denoted as $d^L$, usually applied by a KL divergence loss:
\begin{equation}
\begin{aligned}
    L_{distill} &=L_{logit} =d^L(Y^S,Y^T) \\
    &= \tau^2 KL(softmax(\frac{Y^S}{\tau}),softmax(\frac{Y^T}{\tau}))
\end{aligned}
\label{eq:logit-dis}
\end{equation}
Hyperparameter $\tau$, referred to as temperature, is introduced to put additional control on softening of the signal arising from the output of the teacher network. 
On the other hand, feature distillation utilizes the objective to measure the distance between intermediate features, denoted as $d^F$, as the distillation loss of the student features and teacher features: 
\begin{equation}
    L_{distill}=L_{feature}=d^F (F^S,F^T)
    \label{eq:feature-dis}
\end{equation}
The student is then trained under the integrated objectives: 
\begin{equation}
    L=\alpha L_{task}+\beta L_{distill}
    \label{eq:kd}
\end{equation}
where $\alpha$ and $\beta$ are second hyperparameters controlling the trade-off between the two losses.

\subsection{Feature Alignment: The Missing Piece of Logit-based Distillation}
The primary distinction between the two types of distillation lies in the measurement of the distance between $T$ and $S$. Feature-based methods tend to extract more detailed information from the feature maps, whereas logit-based methods primarily concentrate on the final output. 
The fundamental principle of feature distillation is to achieve feature alignment by defining and reducing the feature distance $L_{feature}=d^F (F^S,F^T)$. Due to the distinct shapes of $T$ and $S$'s intermediate features, it is necessary to convert the features of both entities to a uniform size prior to continuing the alignment procedure. 

There are essentially two categories of alignment methods. As depicted in Fig. \ref{fig2}a, one strategy is to entrust transformation modules, denoted as module $M^T$ and $M^S$ for $T$ and $S$, which facilitates the mapping from the feature space to the latent space. The distance between features can then be determined using the projection of two features:
\begin{equation}
    d^F (F^S,F^T )=\sum d_i^F (M_i^S (F_i^S),M_i^T (F_i^T))
    \label{eq:feature-1}
\end{equation}
where $M_i^S$ and $M_i^T$ are the projection transformations for $F^S_i$ and $F^T_i$, employed in various studies including attention maps \cite{zagoruyko_paying_2017} and factors \cite{kim_paraphrasing_2018}. Despite the decrease in dimensionality, previous studies have reported that the extracted feature representation does indeed result in enhanced performance. However, this highly constructive method may frequently require good priors and careful adjustments. The observation has been made that the process of projecting from a high-dimensional space to a low-dimensional space inherently results in a substantial loss of information \cite{heo_comprehensive_2019}. Several approaches \cite{heo_comprehensive_2019,chen_distilling_2021} have been proposed to enhance transformations by utilizing a more robust prior, as depicted in Fig. \ref{fig2}b, which explicitly defines the mapping function from the student space $F^S$ to the teacher space $F^T$:
\begin{equation}
    d^F (F^S,F^T )=\sum d_i^F (C_i (F_i^S ),F_i^T )
\end{equation}
where $C_i$ is the connector utilized to acquire identical shapes of teacher features. 
Under the current setting, the process of feature alignment demonstrates a remarkable capacity for mitigating information loss; in addition, this method does not require the use of a meticulously crafted prior for feature extraction.
It is crucial to recognize, however, that this method may result in a rigid alignment with task-irrelevant noise in the feature alignment.

\subsection{Implementing Feature Alignment via Logits}
Drawing inspiration from the existing feature alignment methods, we note that 
distinct implementations can be further refined into a unified and straightforward form for feature alignment (Fig. \ref{fig2}c), which is defined as:
\begin{equation}
    d^F (F^S,F^T )=\sum d_i^F (M_i^P (C_i (F_i^S )),M_i^P (F_i^T ))   
    \label{eq:feature-3}
\end{equation}
where $C_i$ is a connector 
to align the number of channels between the student feature map $F^S_i$ and the teacher feature map $F^T_i$. $M^P_i$ is a shared mapping from high-dimensional to low-dimension for both $T$ and $S$.
Currently, the process of feature alignment is executed within the projected space. It is important to highlight that our design aims to achieve the desired outcome of filtering the high-level features, extracting task-related information that is relatively clean, and conducting feature alignment via the "dark knowledge" inside the latent space.
A clearer picture emerges once well-trained teacher blocks are utilized as common projectors:
\begin{equation}
    M_i^P=T_c\circ T_n \circ ... \circ T_{i+1}
\end{equation}
At that time, since 
\begin{equation}
    M_i^P (F_i^T )=T_c\circ T_n \circ ... \circ T_{i+1} (F_i^T )=Y^T
\end{equation}
The objective of feature alignment, $L_{feature}$ in Eq. \ref{eq:feature-3}, becomes:
\begin{equation}
    d^F(F^S,F^T) = \sum d_i^F (M_i^P (C_i (F_i^S )),Y^T)
\end{equation}
Hence, by employing the proposed alignment method, it becomes feasible to execute feature alignment on the teacher's logits without the need for a dedicated definition projector (Fig. \ref{fig2}d). Additionally, leveraging the trained deep blocks from the teacher allows for mitigating the impact of noisy items on the feature alignment to a significant extent. One of the most intriguing aspects of this particular structure is its ability to integrate feature distillation into logit distillation, as we introduce a series of intermediate models, denoted as $N_i$, serving as stepping stones in the overall process with output logits $Y^{N_i}$:
\begin{equation}
    N_i=T_c\circ T_n\circ ... \circ T_{i+1} \circ C_i \circ S_i  \circ ... \circ S_1
\end{equation}
\begin{equation}
Y^{N_i}=N_i(X)
\end{equation}
where $C_i$ follows the pre-definition that connects the intermediate output of $F^S_i$ into the required input size of $F^T_{i+1}$. 
Now, as shown in Fig. \ref{fig2}d, a meticulously designed framework for achieving feature alignment has been developed:
\begin{equation}
\begin{aligned}
    d_i^F (M_i^P (C_i (F_i^S )),Y^T )&=d_i^F (N_i (X),Y^T )\\
    &=d^L (Y^{N_i},Y^T)
\end{aligned}
\end{equation}
The definition leads us to the unified distillation format as the definition for logit distillation (Eq. \ref{eq:logit-dis}) and feature distillation (Eq. \ref{eq:feature-dis}), which apply stepping-stone models to generate intermediate logits for features alignment.  

\begin{figure}[t]
\centering
\includegraphics[width=0.5\textwidth]{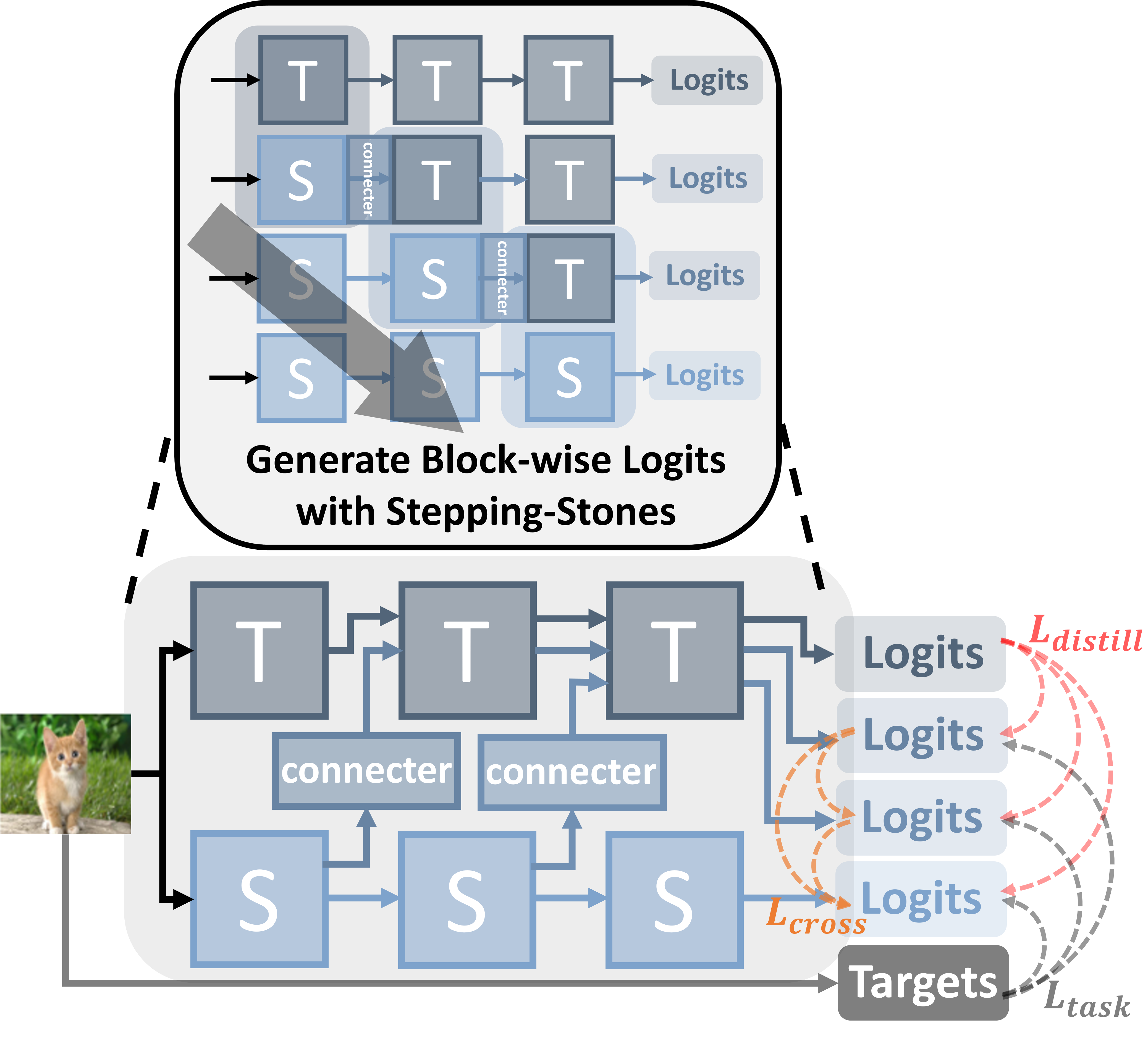} 
\caption{Framework Overview of Block-KD. The stepping stones are executed implicitly within the fundamental dataflow to generate step-by-step logits. The final objectives are defined merely in terms of output logits.}
\label{fig3}
\end{figure}

\subsection{Training Objectives of Overall Framework} 
With the discussed concept, we establish stepping-stones, $N_i$, for feature alignment:
\begin{equation}
    N_i=\{S_1,…,S_i,C_i,T_{i+1},..., T_n,T_c\}
\end{equation}
The training objective for each stepping-stone, $N_i$, is defined to be consistent with the $S$ form:
\begin{equation}
    L^{N_i}=L_{task}^{N_i}+L_{distill}^{N_i}
\end{equation}
where $L_{task}^{N_i}$ denotes the loss incurred by $N_i$ with respect to the task objective, which is computed using cross-entropy loss in the context of classification tasks. $L_{distill}^{N_i}$ refers to the discrepancy between the output logits and the teacher logits in the logit-distillation process.
Currently, the overarching aim of the training is articulated as:
\begin{equation}
\begin{aligned}
    L&=(L_{task}^S+L_{distill}^S)+\sum (L_{task}^{N_i} + L_{distill}^{N_i} )\\
    &=(L_{task}^S+\sum L_{task}^{N_i})+(L_{distill}^S+\sum L_{distill}^{N_i} )\\
    &=L_{task}+L_{distill}
\end{aligned}
\end{equation}
It should be noted that the primary objective of a stepping-stone is to achieve feature alignment.  
Instead of utilizing feature-distillation as described in Eq. \ref{eq:feature-dis}, we propose a revised formulation for $L_{feature}$, as
\begin{equation}
\begin{aligned}
    L_{feature}&=\sum d_i^F (F_i^S,F_i^T)\\
    &= \sum d^L (Y^{N_i},Y^T ) =\sum L_{distill}^{N_i}
\end{aligned}
\end{equation} 
Moreover, as shown in Fig. \ref{fig3}, there is potential for further optimization of the framework with the incorporation of logits distillation between stepping models, donated as $L_{cross}$, which could further enhance the overall effectiveness of the framework. 
We employ an ensemble to streamline its execution:
\begin{equation}
    Y^{ensemble}=\frac{\sum^{i\leq n}_{i=1} Y^{N_i}}{n}
\end{equation}
Then, the process of distillation between stepping stones is described as:
\begin{equation}
    L_{cross}=d^L (Y^S,Y^{ensemble})+ \sum d^L (Y^{N_i},Y^{ensemble})
\end{equation}
All in all, Fig. \ref{fig3} presents the entire process of Block-KD very well. Compared with vallina KD, in addition to obtaining the respective logits of student and teacher, we also obtained the logits of the intermediate state according to the calculation of stepping-stones, and integrated learning objectives with control hyperparameters $\alpha$, $\beta$, $\gamma$:
\begin{equation}
    L=\alpha L_{task}+\beta L_{distill}+\gamma L_{cross}
    \label{final-obj}
\end{equation}

{
\section{Theoretical Analysis}
}
\noindent
{
The improvements brought by the Block-KD framework to the student backbone can be analyzed from various perspectives. 
}


{
First, let us show that distillation through stepping-stones effectively aligns intermediate features. Given two features that need to be aligned, \( F_i^T \) and \( F_i^S \), we denote the difference between them as \( d_i^F(F_i^S, F_i^T) = d^L(Y^{N_i}, Y^T) \) to measure their divergence. We now show how \( d_i^F(F_i^T, F_i^S) \) facilitates the alignment of these features. In this context, consider \( d^L \) to be the KL divergence loss (Eq. \ref{eq:logit-dis}). Regarding the $K$-classes logit, \( Y_k^{N_i} \) of the distilled stepping-stone model and \( Y_k^T \) of the teacher model, where \( k \) represents the class dimension, the gradient of the distillation loss is given by:
\begin{equation}
\frac{\partial d^L}{\partial Y_k^{N_i}} = \frac{1}{\tau} \left( \frac{e^{Y_k^{N_i} / \tau}}{\sum_j e^{Y_j^{N_i} / \tau}} - \frac{e^{Y_k^T / \tau}}{\sum_j e^{Y_j^T / \tau}} \right)
\end{equation}
If the temperature \( \tau \) is high compared to the magnitude of the logits, we can approximate:
\begin{equation}
\frac{\partial d^L}{\partial Y_k^{N_i}} \approx \frac{1}{\tau} \left( \frac{1 + Y_k^{N_i} / \tau}{K + \sum_j Y_j^{N_i} / \tau} - \frac{1 + Y_k^T / \tau}{K + \sum_j Y_j^T / \tau} \right)
\end{equation}
Assuming the logits have been zero-meaned separately for each transfer case such that \( \sum_j Y_j^{N_i} = \sum_j Y_j^T = 0 \), we have:
\begin{equation}
\frac{\partial d^L}{\partial Y_k^{N_i}} \approx \frac{1}{\tau^2 K} (Y_k^{N_i} - Y_k^T)
\end{equation}
At this juncture, let \( C_i(F_i^S) = F_i^P \), given \( Y^{N_i} = M_i^P(F_i^P) \) and \( Y^T = M_i^P(F_i^S) \), we have:
\begin{equation}
\begin{aligned}
\frac{\partial d^L}{\partial F_i^P} &= \sum_k \left(\frac{\partial d^L}{\partial Y_k^{N_i}} \frac{\partial Y_k^{N_i}}{\partial F_i^P}\right) \\
&\approx \frac{1}{\tau^2 K} (Y^{N_i} - Y^T)^\top \frac{\partial M_i^P}{\partial F_i^P} \\
&= \frac{1}{\tau^2 K} (M_i^P(F_i^P) - M_i^P(F_i^T))^\top \frac{\partial M_i^P}{\partial F_i^P}
\end{aligned}
\end{equation}
Then, with a linear approximation, assuming the model \( M_i^P \) is approximately linear near \( F_i^P \), we can retain only the first-order term in Taylor expansion at \( F_i^P \):
\begin{equation}
M_i^P(F_i^T) \approx M_i^P(F_i^P) + \frac{\partial M_i^P}{\partial F_i^P}(F_i^T - F_i^P)
\end{equation}
Substituting this approximation into the original gradient expression, we obtain:
\begin{equation}
\frac{\partial d^L}{\partial F_i^P} \approx \frac{1}{\tau^2 K} \left(\frac{\partial M_i^P}{\partial F_i^P}(F_i^P - F_i^T)\right)^\top \frac{\partial M_i^P}{\partial F_i^P}
\end{equation}
This gradient directly influences the alignment of \( F_i^P \) and \( F_i^S \). In the gradient descent method, the direction of adjustment for \( F_i^P \) should be toward reducing the difference between \( F_i^P \) and \( F_i^S \), and this is determined by the direction and magnitude of \( \frac{\partial M_i^P}{\partial F_i^P} \).
}

\begin{table*}[t]
\centering
\caption{\textbf{Results of top-1 accuracy (\%) on CIFAR-100 under uniform architecture.} Block[KD], Block[DKD], and Block[MLKD] refer to proposed block-wise logit distillation under the logit objectives of classical KD\cite{hinton_distilling_2015}, DKD\cite{zhao_decoupled_2022}, and MLKD\cite{jin2023multi} respectively. Block[DKD]$^\dag$ indicates the light version, utilizing the last two stepping models. $^*$ signifies configurations that include data augmentation and 480 training epochs following MLKD\cite{jin2023multi}. All results are the average over 5 trials.}
\resizebox{\textwidth}{!}{
\begin{tabular}{cc|cccccc}
\toprule
\multirow{4}{*}[-0.30ex]{Method} 
& \multirow{2}{*}[-0.35ex]{Teacher} & ResNet56 & ResNet110 & ResNet32×4 & WRN40-2 & WRN40-2 & VGG13 \\
&      & 72.34    & 74.31     & 79.42      & 75.61    & 75.61    & 74.64 \\
& \multirow{2}{*}[0.35ex]{Student}  & ResNet20  & ResNet32   & ResNet8×4   & WRN16-2  & WRN40-1 & VGG8  \\
  &                          & 69.06    & 71.14     & 72.50      & 73.26    & 71.98    & 70.36 \\
\midrule
\multirow{6}{*}{Feature}                                                      & FitNet   \cite{romero_fitnets_2015}              & 69.21    & 71.06     & 73.50      & 73.58    & 72.24    & 71.02 \\
   & RKD  \cite{park_relational_2019}   & 69.61    & 71.82     & 71.90      & 73.35    & 72.22    & 71.48 \\
   & CRD  \cite{tian_contrastive_2022}    & 71.16    & 73.48     & 75.51      & 75.48    & 74.14    & 73.94 \\
   & OFD  \cite{heo_comprehensive_2019}  & 70.98    & 73.23     & 74.95      & 75.24    & 74.33    & 73.95 \\
   & ReviewKD  \cite{chen_distilling_2021}  & 71.89    & 73.89     & 75.63      & 76.12    & 75.09    & 74.84 \\
   &FAM-KD \cite{pham2024frequency} &72.15&74.45& 76.84 & 76.47 & 75.40 & / \\
\midrule

\multirow{8}{*}[-9.5ex]{w/ Logit}                                               
&DML \cite{zhang_deep_2018} &69.52 &72.03 &72.12 &73.58 &72.68 &71.79 \\
&TAKD \cite{mirzadeh_improved_2019} &70.83 &73.37 &73.81 &75.12 &73.78 &73.23 \\

& WSLD \cite{zhourethinking} &  72.15 &  74.12 &  76.05 & / & 74.48 & / \\
& WCoRD \cite{chen2021wasserstein} & 71.92& 74.20& 76.15& 76.11 & 74.72 & / \\
& NKD \cite{yang2023knowledge} & / & / & 76.35 & / & / & 74.86 \\
& CKD \cite{zhang2024collaborative} &72.03& 74.14 & 76.62 & 76.29 & 74.85  & 74.86 \\

\cmidrule{2-8}
& KD \cite{hinton_distilling_2015}  & 70.66    & 73.08     & 73.33      & 74.92    & 73.54    & 72.98 \\
& Block[KD]   & 71.71        & 73.95         & 76.26          & 75.67        & 74.83        & 74.54    \\

\cmidrule{2-8}
& DKD \cite{zhao_decoupled_2022} & 71.97    & 74.11     & 76.32      & 76.24    & 74.81    & 74.68 \\
& OFD+DKD &71.12 &73.53 &75.22 &76.03 &74.73 &74.73 \\
& ReviewKD+DKD &71.63 &\textbf{74.20} &76.01 &76.15 &74.86 &74.96 \\
& Block[DKD]$^\dag$ & \textbf{72.06} &74.15 &77.18 &76.26 &75.14 &\textbf{75.06}\\
& Block[DKD]  $ $   &72.03  &74.17   & \textbf{77.28}    &\textbf{76.30}     &\textbf{75.21}    &75.05  \\

\cmidrule{2-8}
& MLKD$^*$ \cite{jin2023multi} &72.19 &74.11 &77.08 &76.63 &75.35 &75.18  \\
& Block[MLKD]$^*$ &\textbf{72.43} &\textbf{74.72} &\textbf{77.84} &\textbf{76.76} &\textbf{75.97} &\textbf{75.34} \\

\bottomrule
\end{tabular}}
\label{tab1}
\end{table*}

{Besides,} the stepping models constructed during training provide a moderately performing model akin to an implicit teacher-assistance \cite{mirzadeh_improved_2019} mechanism. The theoretical analysis of performance enhancement by creating these intermediate models is discussed in \cite{mirzadeh_improved_2019}. This analysis is primarily based on VC theory \cite{vapnik1999overview_cp} and its relevant applications in the field of knowledge distillation \cite{lopez2015unifying_cp}. 
{
Consider an initial scenario wherein a student begins learning without prior knowledge. According to the principles outlined in VC theory \cite{vapnik1999overview_cp}, the classification error of a classifier $f_s$ is decomposable in the following manner:
\begin{equation}
    R(f_s) - R(f_r) \leq O\left(\frac{|F_s|_C}{n^{\alpha_{sr}}}\right) + \epsilon_{sr}
\end{equation}
Here, $O(\cdot)$ symbolizes the estimation error associated with the statistical learning processes given a specified dataset size, while $\epsilon_{sr}$ delineates the approximation error reflective of the model's inherent capability. Within this framework, $f_r \in F_r$ represents the authentic (ground truth) target function, and $f_s \in F_s$ denotes the learner or student function. $R$ specifies the error magnitude, $|\cdot|_C$ quantifies the function class capacity, $n$ denotes the dataset size, and $\alpha_{sr}$ ranges between 1/2 to 1, indicating a learning rate which approaches 1/2 for more complex tasks and nears 1 for simpler tasks.
Similarly, assuming $f_t \in F_t$ as the teacher function, we have:
\begin{equation}
    R(f_t) - R(f_r) \leq O\left(\frac{|F_t|_C}{n^{\alpha_{tr}}}\right) + \epsilon_{tr}
\end{equation}
where $\alpha_{tr}$ and $\epsilon_{tr}$ are correspondingly defined to encapsulate the learning dynamics when the teacher is educated from scratch.
Building upon the groundwork established by \cite{lopez2015unifying_cp}, and assuming that training is conducted through pure distillation ($\lambda = 1$), we facilitate direct knowledge transfer from the teacher to the student, thereby establishing a baseline for knowledge distillation. From this baseline, we derive:
\begin{equation}
    R(f_s) - R(f_t) \leq O\left(\frac{|F_s|_C}{n^{\alpha_{st}}}\right) + \epsilon_{st}
\end{equation}
in which $\alpha_{st}$ and $\epsilon_{st}$ relate to the student's learning process from the teacher. \cite{lopez2015unifying_cp} highlighted the necessity for $|F_t|_C$ to remain minimal to ensure that traditional knowledge distillation surpasses rudimentary learning processes. Furthermore, it is imperative to recognize that $\alpha_{sr} \leq \alpha_{st}$ and $\alpha_{sr} \leq \alpha_{tr}$, with a larger gap indicating a reduced rate of learning. Additionally, according to assumptions made by \cite{hinton_distilling_2015}, $\epsilon_{tr} + \epsilon_{st} \leq \epsilon_{sr}$. This setup confirms effective knowledge distillation:
\begin{equation}
    O\left(\frac{|F_t|_C}{n^{\alpha_{tr}}} + \frac{|F_s|_C}{n^{\alpha_{st}}}\right) + \epsilon_{tr} + \epsilon_{st} \leq O\left(\frac{|F_s|_C}{n^{\alpha_{sr}}}\right) + \epsilon_{sr}
\end{equation}
This relationship suggests that the aggregate error bound in traditional distillation is less than or equal to that in scratch learning, valid asymptotically as $n \rightarrow \infty$. In scenarios where the sample size is finite and $|F_t|_C$ is excessively large, these inequalities might fail, indicating potential shortcomings in the distillation process. Another failure scenario for traditional knowledge distillation arises when there is a significant disparity in capacity between the student and teacher, particularly when $\alpha_{st}$ is minimal.
Leveraging their insights and analytical framework, let us introduce an intermediate stepping-stone model, donated as $f_n$, between the student and teacher:
\begin{equation}
    R(f_s) - R(f_n) \leq O\left(\frac{|F_s|_C}{n^{\alpha_{sn}}}\right) + \epsilon_{sn}
\end{equation}
and subsequently, this stepping-stone model assimilates knowledge from the teacher:
\begin{equation}
    R(f_n) - R(f_t) \leq O\left(\frac{|F_n|_C}{n^{\alpha_{nt}}}\right) + \epsilon_{nt}
\end{equation}
Here, $\alpha_{sn}, \epsilon_{sn}, \alpha_{nt},$ and $\epsilon_{nt}$ are defined to elucidate the learning dynamics within this tripartite framework. It is pertinent to note that the incorporation of an intermediary simplifies the learning process, whether it is the student learning from the intermediary or the intermediary from the teacher, i.e., $
\alpha_{st} \leq \alpha_{sn} \text{ and } \alpha_{st} \leq \alpha_{nt}
$, thus,
\begin{equation}
O\left(\frac{|F_n|_C}{n^{\alpha_{nt}}} + \frac{|F_s|_C}{n^{\alpha_{sn}}}\right) \leq O\left(\frac{|F_s|_C}{n^{\alpha_{st}}}\right)
\end{equation}
Additionally, under the assumptions made by \cite{hinton_distilling_2015}, 
$\epsilon_{nt} + \epsilon_{sn} \leq \epsilon_{st}$, we have,
\begin{equation}
O\left(\frac{|F_n|_C}{n^{\alpha_{nt}}} + \frac{|F_s|_C}{n^{\alpha_{sn}}}\right) + \epsilon_{nt} + \epsilon_{sn} 
\leq O\left(\frac{|F_t|_C}{n^{\alpha_{tr}}} + \frac{|F_s|_C}{n^{\alpha_{st}}}\right) + \epsilon_{st},
\end{equation}
These relationships corroborate that the upper bound of error with a stepping-stone model is smaller than in vanilla knowledge distillation. 
In scenarios characterized by significant performance disparities between the student and the teacher, the integration of a stepping-stone model intermediary effectively partitions the modest $\alpha_{st}$ into two more tractable components, $\alpha_{sn}$ and $\alpha_{nt}$. This strategic division substantially enhances the efficiency of the knowledge distillation process, optimizing the learning trajectory for the student.
}

Moreover, the principles underlying Block-KD's branch structure are reminiscent of self-distillation, and its benefits can be understood through a multi-view hypothesis \cite{allen2020towards_cp}.
Introducing $L_{cross}$, which averages votes during the training process, often leads to more effective teaching labels. This approach aligns with constructing an output ensemble \cite{wang_knowledge_2022}. During training, it can yield performance that surpasses even the initial teacher model, thereby elevating the overall framework's upper bound. These observations generally align with empirical experiments \cite{wang_knowledge_2022}.

\section{Experiments}


\begin{table*}[tb]
\centering
\caption{\textbf{Results of top-1 accuracy (\%) on CIFAR-100 under non-uniform architecture.} Block[KD], Block[DKD], and Block[MLKD] refer to proposed block-wise logit distillation under the logit objectives of classical KD\cite{hinton_distilling_2015}, DKD\cite{zhao_decoupled_2022}, and MLKD\cite{jin2023multi} respectively. Block[DKD]$^\dag$ indicates the light version, utilizing the last two stepping models. $^*$ signifies configurations that include data augmentation and 480 training epochs following MLKD\cite{jin2023multi}. All results are the average over 5 trials.}
\resizebox{\textwidth}{!}{
\begin{tabular}{cc|ccccc}
\toprule
\multirow{4}{*}[-0.30ex]{Method}  & \multirow{2}{*}[-0.35ex]{Teacher} & ResNet32×4    & WRN40-2 & VGG13  & ResNet50 & ResNet32×4 \\
&      & 79.42    & 75.61     & 74.64      & 79.34    & 79.42 \\
& \multirow{2}{*}[0.35ex]{Student} &ShuffleNetV1 &ShuffleNetV1 &MobileNetV2 &MobileNetV2 &ShuffleNetV2\\
  &                          &70.50& 70.50 &64.60 &64.60 &71.82 \\
  
\midrule
\multirow{6}{*}{Feature}                                                      & FitNet \cite{romero_fitnets_2015}                  &73.59 &73.73 &64.14 &63.16 &73.54 \\
   & RKD \cite{park_relational_2019}      &72.28 &72.21 &64.52 &64.43 &73.21 \\
   & CRD \cite{tian_contrastive_2022}      &75.11 &76.05 &69.73 &69.11 &75.65 \\
   & OFD \cite{heo_comprehensive_2019}    &75.98 &75.85 &69.48 &69.04 &76.82 \\
   & ReviewKD \cite{chen_distilling_2021}    &77.45 &77.14 &70.37 &69.89 &77.78 \\
   &FAM-KD \cite{pham2024frequency} &77.76&77.57& 70.88 & / & 78.41  \\
\midrule

\multirow{8}{*}[-9.5ex]{w/ Logit}                                                       
&DML  \cite{zhang_deep_2018} &72.89 &72.76 &65.63 &65.71 &73.45  \\
&TAKD \cite{mirzadeh_improved_2019} &74.53 &75.34 &67.91 &68.02 &74.82  \\

& WSLD \cite{zhourethinking} & 75.46 & 76.21 &  / & / & 75.93  \\
& WCoRD \cite{chen2021wasserstein} &75.77&76.68&70.02&76.48&/\\
& NKD \cite{yang2023knowledge} &75.31 &75.96& 69.39 &68.72 &76.26 \\



\cmidrule{2-7}
& KD \cite{hinton_distilling_2015} &74.07 &74.83 &67.37 &67.35 &74.45 \\
& SDD+DKD \cite{luo2024scale} &76.30&76.65& 68.79&69.55&76.67\\
& Block[KD]   & 77.51        &   76.82      & 69.33         & 71.42        & 77.52     \\

\cmidrule{2-7}
& DKD \cite{zhao_decoupled_2022}  &76.45 &76.70 &69.71 &70.35 &77.07 \\
& OFD+DKD &76.99 &76.85 &69.72 &70.23 &77.34 \\
& ReviewKD+DKD &77.43 &77.23 &70.23&70.44 &77.52 \\
& SDD+DKD \cite{luo2024scale} &77.30&77.21& \textbf{70.25}&71.36&78.05\\
& Block[DKD]$^\dag$ & 77.88 & 77.21 & 69.77 & 72.19 & 78.07\\
& Block[DKD]  $ $   &\textbf{78.12}  &\textbf{77.26}   & 69.82    &\textbf{72.24}     &\textbf{78.35} \\

\cmidrule{2-7}
& MLKD$^*$ \cite{jin2023multi} &77.18 &77.44 &70.57 &71.04 &78.44  \\
& Block[MLKD]$^*$  &\textbf{78.29} &\textbf{77.86} &\textbf{70.88} &\textbf{72.54} &\textbf{79.00}\\

\bottomrule
\end{tabular}
}
\label{tab2}
\end{table*}

\subsection{Experiments Details on Visual Tasks}

\subsubsection{Datasets}
Three widely researched datasets are used in experiments: (1) CIFAR-100 \cite{krizhevsky2009learning}, the well-known classification dataset of 32×32 images, with 50K training samples and 10K test samples for 100 classes; (2) ImageNet \cite{russakovsky_imagenet_2015}, the challenging large-scale classification dataset of high-resolution images, with 1.28M training samples and 50K test samples for 1000 classes; (3) MS-COCO \cite{lin_microsoft_2014}, the general object detection dataset, with 118,000 training samples and 5,000 validation samples from 80 categories. 
\subsubsection{Setting}
In our experiments, two primary conditions are addressed. 1) Uniform case in which both the teacher and student models have the same architectural design. 2) Non-uniform case in which the structures of the two models are distinct. We employ a variety of network architectures, including ResNet \cite{he_deep_2016}, WRN \cite{zagoruyko_wide_2017}, VGG \cite{simonyan_very_2015}, ShuffleNet-V1/V2 \cite{zhang_shufflenet_2018,ma_shufflenet_2018}, and MobileNet-V1/V2 \cite{howard_mobilenets_2017,sandler_mobilenetv2_2018}. 
We report the typical or recent methods of both feature and logit distillation for comparison: FitNet \cite{romero_fitnets_2015}, RKD  \cite{park_relational_2019}, CRD \cite{tian_contrastive_2022}, OFD \cite{heo_comprehensive_2019}, ReviewKD \cite{chen_distilling_2021}, vallina KD \cite{hinton_distilling_2015},
DML \cite{zhang_deep_2018}, TAKD \cite{mirzadeh_improved_2019}, DKD \cite{zhao_decoupled_2022}, MLKD \cite{jin2023multi}.

\subsubsection{Implementation Details}
Firstly, the positions of the feature-level distillation need to be determined. The term "block" refers to the traversed layers between distillation positions. We follow the standard block strategy of feature distillation \cite{heo_comprehensive_2019,chen_distilling_2021,wang_knowledge_2022}, which typically involves dividing blocks based on their resolution prior to downsampling, consistent with the practice of convolutional architecture (e.g. 3-blocks or 4-blocks ResNet architecture). The objectives of the stepping stone, $\{L^{N_i}\}_{i \leq n}$, are divided by the coefficient $2^{n-i}$, where $n$ is the total number of blocks, in order to reconcile the correlation between step $L^N$ and final $L^S$, thereby preventing overfitting on stepping-stones. 
Our framework employs the same connectors as OFD \cite{heo_comprehensive_2019}, which consist of a single 1×1 convolutional layer coupled with a batch normalization layer for channel combination. Then, our experimental implementation adheres to the standard strategy identified in previous works \cite{chen_distilling_2021,tian_contrastive_2022,zhao_decoupled_2022}, specifically with regard to each dataset:

\noindent
\textbf{On CIFAR-100}. Models are trained using SGD for 240 epochs with batch size of 64.  The weight decay and momentum are set to 5e-4 and 0.90, respectively. The learning rates for VGG, ResNet, and WRN students are 0.05, while those for ShuffleNet and MobileNet are 0.01. Using a multi-step schedule, the learning rate is divided by 10 at the 150, 180, and 210 epochs. 
The steady weight of all objectives, $L_{task}$, $L_{distill}$, and $L_{cross}$, is set to 1.0, and a 20-epoch linear warmup is used for $L_{distill}$ and $L_{cross}$.  The hyperparameters within the logit distance, $d^L$, of both classical KD and DKD are set according to DKD's original paper \cite{zhao_decoupled_2022}, in which the temperature is set to 4 and the controlling factors on various architectures are altered.

\noindent
\textbf{On ImageNet.} Following \cite{chen_distilling_2021,tian_contrastive_2022,zhao_decoupled_2022} to the letter, we train the models for 100 epochs with the batch size of 512, the learning rate of 0.2 divided by 10 every 30 epochs, and the weight decay of 1e-4. We apply the same configuration of DKD loss \cite{zhao_decoupled_2022}, in which the inside hyperparameters of DKD, $\alpha_{DKD}$ and $\beta_{DKD}$, are set to 0.5, 0.5 for the uniform pair of ResNet-34 and ResNet-18 and to 0.5, 2.0 for the non-uniform pair of ResNet-50 and MobileNet-V1. The steady weight of all objectives, $L_{task}$, $L_{distill}$, and $L_{cross}$, is set to 1.0 and a linear warmup is utilized for the adaptation of stepping-stones in the first 5 epochs.

\noindent
\textbf{On MS-COCO.} Implementation follows the settings in \cite{chen_distilling_2021}. We employ the two-stage method, Faster RCNN with FPN \cite{ren_faster_2015,lin_feature_2017} as the foundation. All students are trained with the 1x scheduler (using the identical schedulers and task-specific loss weights as in Detectron2 \cite{wu2019detectron2}). The backbone and FPN are viewed as a single complete block, and the connector used to transfer FPN output features employs ABF structure proposed in \cite{chen_distilling_2021}. $L_{distill}$ and $L_{cross}$ are defined on DKD with the same setting in \cite{zhao_decoupled_2022}.

\noindent
All results of compared methods are reported in their original papers or reproduced by previous works \cite{chen_distilling_2021,tian_contrastive_2022,zhao_decoupled_2022}.

{
\subsection{Experiments Details on NLP tasks}
}
\noindent
{
\subsubsection{Datasets}
We have also extended the application of Block-KD to evaluate its performance across various benchmarks within the field of NLP: Microsoft Research Paraphrase Matching (MRPC) \cite{dolan2005automatically} and Quora Question Pairs (QQP) \cite{sharma2019natural} are used for assessing paraphrase similarity; the Recognizing Textual Entailment (RTE) \cite{bentivogli2009fifth} task for natural language inference; and the Corpus of Linguistic Acceptability (CoLA) \cite{warstadt2019neural} for determining linguistic acceptability. Following \cite{wang2018glue}, we use classification accuracy as the evaluation metric for the RTE datasets, both the accuracy and the F1 score for MRPC and QQP, and the Matthew’s correlation coefficient for CoLA.
}

{
\subsubsection{Settings}
For our experiments on NLP benchmarks, which entail sentence or sentence-pair classification tasks, we leverage the architecture from the original BERT model \cite{devlin2018bert}. 
The standard configuration of the teacher model, referred to as BERT\textsubscript{12}, includes 12 transformer layers, a hidden size of \(d=768\), a feedforward size of \(d_i=3072\), and $h=12$ attention heads, totaling $109$ million parameters. For the student models, we use BERT\textsubscript{4} (with \(d=312\), \(d_i=1200\), $h=12$, and $14.5$ million parameters) and BERT\textsubscript{6} (with \(d=768\), \(d_i=3072\), $h=12$, and $67.0$ million parameters). 
To enable effective knowledge distillation, the teacher model is pre-finetuned for each task \cite{sun2019patient}, and the student models are initialized with a generalized distillation strategy derived from \cite{jiao2019tinybert, li2020bert}.
}

{
\subsubsection{Implementation Details}
During the distillation phase, the teacher model supplies logits as soft labels for the students, employing a standard KL divergence loss (Eq. \ref{eq:logit-dis}) for distillation. Training involves 10 epochs of fine-tuning with a batch size of 32, a learning rate of \(2 \times 10^{-5}\), and distillation coefficients \(\alpha=0.2\), \(\beta=0.8\) in Eq. \ref{eq:kd}, with \(\tau=1.0\) in Eq. \ref{eq:logit-dis} for tasks of CoLA, MRPC, and QQP. For the RTE task, coefficients are adjusted to \(\alpha=\beta=0.5\) and \(\tau=3.0\). 
Within the Block-KD framework, for simplicity, BERT’s embeddings and encoder are treated as a single block and the classifier as a replaceable block, facilitating a lightweight framework verification test. At this stage, only one stepping-stone is created, merging the student’s embeddings and encoder with the teacher’s classifier, with a linear connector placed between the student’s encoder and the teacher’s classifier. In this simplified setup, $L_{cross}$ is not activated.
}

\begin{table}[tb]
    \centering
    \caption{\textbf{Results of top-1 and top-5 accuracy (\%) on the ImageNet validation.} Teacher-student pairs are ResNet-34/ResNet-18 and ResNet-50/MobileNet-V1 for uniform and non-uniform architecture, respectively.
    }
    \resizebox{0.5\textwidth}{!}{
    \begin{tabular}{cc|cc|cc}
    \toprule
         \multicolumn{2}{c|}{} &Top-1 &Top-5  & Top-1 & Top-5  \\
         \midrule
         \multirow{4}{*}{Method} &\multirow{2}{*}{Teacher} &\multicolumn{2}{c|}{ResNet34} &   \multicolumn{2}{c}{ResNet50} \\
         & &73.31 &91.42 &76.16 &92.86\\
         &\multirow{2}{*}[0.2ex]{Student} &\multicolumn{2}{c|}{ResNet18} &   \multicolumn{2}{c}{MobileNet-V2} \\
         & &69.75 &89.07  &68.87 &88.76 \\
         \midrule
         \multirow{6}{*}{Feature} &AT \cite{zagoruyko_paying_2017} &70.69 &90.01&69.56&89.33 \\
         &OFD \cite{heo_comprehensive_2019} &70.81 &89.98 &71.25 &90.34 \\
         &CRD \cite{tian_contrastive_2022} &71.17 &90.13 &71.37 &90.41 \\
         &WCoRD \cite{chen2021wasserstein} & 71.49 & 90.16 & / & / \\
         &ReviewKD \cite{chen_distilling_2021} &71.61 &90.51 &72.56 &91.00 \\
         &MGD \cite{yang2022masked} & 71.58 & 90.35 & 72.35 & 90.71 \\

         \midrule
          \multirow{5}{*}{Logit} 
          &KD \cite{hinton_distilling_2015} &71.03 &90.05 &70.50 &89.80  \\
          &DML \cite{zhang_deep_2018} &70.82 &90.02 &71.35 &90.31 \\
          &WSLD \cite{zhourethinking} & 71.54 & 90.25 & 72.02 & 90.70 \\
            &DKD \cite{zhao_decoupled_2022} &71.70 &90.41 &72.05 &91.05   \\
         &NKD \cite{yang2023knowledge} & 71.96 & / & 72.58 & / \\
            
        \midrule
          \multirow{4}{*}{Feature + Logit} 
        &SRRL \cite{yang2021knowledge} & 71.73 & 90.60 & 72.49 & 90.92 \\
        &MGD+WSLD \cite{yang2022masked} & 71.80 & 90.40 & 72.59 & 90.94\\
        & CKD \cite{zhang2024collaborative} &71.79&90.58& /&/ \\
        &Block[DKD] &\textbf{72.26} &\textbf{90.76} &\textbf{73.11} &\textbf{91.33}  \\
    \bottomrule
                  
    \end{tabular}
    }
    \label{tab3}
\end{table}

\begin{table*}[tb]
    \centering
    \caption{\textbf{Results on MS-COCO based on Faster-RCNN-FPN with AP evaluated on val2017.} Teacher-student pairs are ResNet-101 (R-101) \& ResNet-18 (R-18), ResNet-101 \& ResNet-50 (R-50) and ResNet-50 \& MobileNet-V2 (MV2).}
    \resizebox{\textwidth}{!}{
    \begin{tabular}{cc|ccc|ccc|ccc}
    \toprule
  \multicolumn{2}{c|}{}  &\multicolumn{3}{c|}{R101 \& R-18} &\multicolumn{3}{c|}{R101 \& R-50} &\multicolumn{3}{c}{R50 \& R-MV2} \\
   \multicolumn{2}{c|}{}    & AP &AP$_{50}$ &AP$_{75}$ & AP &AP$_{50}$ &AP$_{75}$ & AP &AP$_{50}$ &AP$_{75}$ \\
    \midrule
  \multirow{2}{*}{Method} &teacher &42.04 &62.48 &45.88 &42.04 &62.48 &45.88 &40.22 &61.02 &43.81 \\
    &student &33.26 &53.61 &35.26 &37.93 &58.84 &41.05 &29.47 &48.87 &30.90 \\
    \midrule 
\multirow{3}{*}{Feature}    &FitNet \cite{romero_fitnets_2015} &34.13 &54.16 &36.71 &38.76 &59.62 &41.80 &30.20 &49.80 &31.69   \\
    
    &FGFI \cite{wang2019distilling} &35.44 &55.51 &38.17 &39.44 &60.27 &43.04 &31.16 &50.68 &32.92    \\
     &ReviewKD \cite{chen_distilling_2021}  &\textbf{36.75} &56.72 &34.00 &\textbf{40.36} &60.97 &\textbf{44.08} &33.71 &53.15 &36.13    \\
    \midrule
 \multirow{3}{*}[-0.65ex]{Logit}   &KD \cite{hinton_distilling_2015} &33.97 &54.66 &36.62 &38.35 &59.41 &41.71 &30.13 &50.28 &31.35  \\
  
    &DKD \cite{zhao_decoupled_2022} &35.05 &56.60 &37.54 &39.25 &60.90 &42.73 &32.34 &53.77 &34.01  \\
    \cmidrule{2-11}
    &Block[DKD]   &36.32  & \textbf{57.30}   & \textbf{38.97}   &40.34   & \textbf{61.47}   & 43.79   & \textbf{34.45}   & \textbf{54.76}   &\textbf{36.82} \\
    \bottomrule
    \end{tabular}
    }
\end{table*}

\subsection{Main Results}
\noindent
\textbf{Results on CIFAR-100 image classification.} Tab. \ref{tab1} and Tab. \ref{tab2} show the performance of various distillation schemes under both uniform and non-uniform architectures. To implement the objectives of Block-KD (Eq. \ref{final-obj}), we employ the logit-based distance ($d^L$) suggested by vallina KD and DKD, respectively. Notably, with the Block-KD framework, the logit-based scheme is capable of achieving consistent development. The Block-KD based on vallina KD has been able to outperform complex feature methods in the majority of instances, and the scheme with DKD has yielded the best results in nearly all cases. On this basis, the lightweight version with only the last two stepping models reduces computational costs to levels comparable to OFD and ReviewKD for comparison (see Block[DKD]$^\dag$ in Tab. \ref{tab1} and Tab. \ref{tab2}). 
We also conduct experiments with combinations of feature-based OFD, ReviewKD, and logit-based DKD in Tab. \ref{tab1} and Tab. \ref{tab2}. 
Notably, the results of combined methods tend to fall between the individual performance levels of feature and logit methods. This suggests that the training objectives of logits and features may diverge, a phenomenon consistent with related experiments in 
\cite{tian_contrastive_2022}. While in specific limited application scenarios, block-KD delivers stable performance improvements over logit methods, which underscores the advantages of block-KD compared to feature methods.

\noindent
\textbf{Results on ImageNet image classification.} Tab. \ref{tab3} illustrates the distillation results on the large-scale task with both Top-1 and Top-5 accuracies, demonstrating that our method still outperforms the state-of-the-art distillation methods on such a large and complex dataset. Notably, the logit distillation method DKD and the feature distillation method ReviewKD have advantages in different indices, whereas our proposed block-wise logit distillation has completely surpassed them.

\noindent
\textbf{Results on MS-COCO object detection.} To verify the generality of the Block-KD method, we applied it to the object detection task. Faster-RCNN-FPN \cite{ren_faster_2015,lin_feature_2017} serves as the foundation, while AP, AP50, and AP75 are used for the evaluation metric. The Block-KD framework needs to be extended to match the FPN structure in this case.  The backbone and FPN are seen as one complete block.  The connecter is used to transfer FPN output features from the student to the teacher. The teacher ROI generates intermediate logits and instructed logits based on the proposals from the student RPN module, using both student-transferred features and raw features. The distillation goal of Block-KD can then be defined using those step logits and the student's ROI output.  Since the output of FPN is a compilation of several features, which limits the use of the simple convolution structure as in classification tasks, the ABF proposed in ReviewKD \cite{chen_distilling_2021} is employed as the connector portion for feature alignment.

\begin{table}[tb]
    \centering
    \caption{\textbf{Results on NLP tasks with BERT models.} The models are evaluated on datasets of MRPC, QQP, RTE, and CoLA, respectively.}
    \resizebox{0.48\textwidth}{!}{
    \begin{tabular}{l|c|c|c|c}
    \toprule
        & \begin{tabular}[c]{@{}c@{}}MRPC \\ {(Acc/F1)}\end{tabular} 
        & \begin{tabular}[c]{@{}c@{}}QQP \\ {(Acc/F1)}\end{tabular} 
        & \begin{tabular}[c]{@{}c@{}}RTE \\ {(Acc)}\end{tabular} 
        & \begin{tabular}[c]{@{}c@{}}CoLA \\ {(Correlation)}\end{tabular} \\   
        \midrule
        BERT\textsubscript{4}  &82.11/87.69&86.97/82.73&65.70&18.50 \\
        w/ vanilla KD         & 85.05/89.54   & 87.82/83.85  & 66.43     & 20.55  \\
        w/ Block-KD  & \textbf{86.52/90.73}   & \textbf{89.20/85.47}  & \textbf{68.23}     & \textbf{22.65}  \\
        \midrule
        BERT\textsubscript{6} &86.03/90.09&89.54/85.93&71.84&20.72\\
         w/ vanilla KD     & 87.01/90.94   & 89.95/86.39  & 72.20     & 41.30  \\
          w/ Block-KD  & \textbf{87.25/91.13}   & \textbf{90.56/87.30}  & \textbf{73.29}     & \textbf{45.04}  \\
        \bottomrule
    \end{tabular}}
    \label{tab:bert}
\end{table}

\noindent
{\textbf{Results for BERT Models on NLP Tasks.} Tab. \ref{tab:bert} 
presents the distillation outcomes for BERT models across four NLP tasks. Unlike in visual tasks, the Block-KD framework incorporates only one intermediary stepping-stone model in this context, and involves minimal additional overhead—just the teacher's classifier and a linear connector. This streamlined setup is designed to validate the method's generalizability. Observations indicate that employing a distillation loss, alongside training the student model with hard targets, can moderately enhance performance. Furthermore, the Block-KD framework seems to amplify these benefits, enhancing the efficacy of the distillation process. Consequently, our method remains well-suited for transformer architectures and is effective across various applications within the NLP domain, showcasing the framework's adaptability and broad applicability.
}

\begin{table}[tb]
    \centering
    \caption{\textbf{Comparison of different implementations in feature alignment.} ResNet32×4 and ResNet8×4 are paired on CIFAR-100.
    The experimental results are determined under two conditions: 1) Only feature-based objective, $L_{feature}$ is used to achieve feature alignment, in accordance with the schemes proposed in each work. 2) Based on $L_{feature}$, additionally define a logit-based objective, $L_{logit}$, that follows the standard KD loss (Eq. \ref{eq:logit-dis}) for output logits.}
    \resizebox{0.47\textwidth}{!}{
    \begin{tabular}{c|c|c|c}
    \toprule
   \multirow{2}{*}{Mechanism} &\multirow{2}{*}{Method} &\multicolumn{2}{c}{Top-1 Acc (\%)}  \\
   \cline{3-4}
        &  &$L_{feature}$ & w/ $L_{logit}$  \\
         \midrule
         None &Baseline &72.50 &73.33 \\
         \midrule
         \multirow{3}{*}{\begin{tabular}[c]{@{}c@{}}Alignment via \\ Projected Features \\
         \end{tabular}} &AT \cite{zagoruyko_paying_2017} &73.44 &74.53 \\
         &FT \cite{kim_paraphrasing_2018} &72.86 &74.62 \\
         &VID \cite{ahn2019variational} &73.09 &74.56 \\
         \midrule
         \multirow{3}{*}{\begin{tabular}[c]{@{}c@{}} Direct Alignment\\ on Raw Features\end{tabular}} &FitNets \cite{romero_fitnets_2015} &73.50 &74.66 \\
         &AB \cite{heo_knowledge_2019} &73.17 &74.40 \\
         &OFD \cite{heo_comprehensive_2019} &74.95 &75.27 \\ 
         \midrule
         \multicolumn{1}{c|}{\begin{tabular}[c]{@{}c@{}}
              Alignment via \\ Block-wise Logits
         \end{tabular}} &Block &\textbf{75.41} &\textbf{75.94} \\
         \bottomrule    
    \end{tabular}}`
    \label{tab:feature-align}
\end{table}

\subsection{Analysis and Discussion}

\begin{table}[tb]
    \centering
    \caption{\textbf{Ablation study of training objectives (Eq. \ref{final-obj}).}  ResNet32×4 and ResNet8×4 are paired on CIFAR-100. }
    \resizebox{0.47\textwidth}{!}{
    \begin{tabular}{cccccc}
    \toprule
       \multicolumn{2}{c|}{$L^{S}$} &\multicolumn{3}{c|}{$L^{N}$} & \multirow{2}{*}[-0.5ex]{\begin{tabular}[c]{@{}c@{}}
            Top-1 \\ (\%)
       \end{tabular}} \\
       \cmidrule{1-5}
       \multicolumn{1}{c}{$L_{task}^{S}$}  &\multicolumn{1}{c|}{$L_{distill}^{S}$}  &\multicolumn{1}{c}{$L_{task}^{N}$}  &\multicolumn{1}{c}{$L_{distill}^{N}$}  &\multicolumn{1}{c|}{$L_{cross}$}  &  \\
       \midrule
       \ding{51}   & & & & &72.50 \\
       \ding{51}   &\ding{51} & & & &73.33 \\
       \ding{51}   & &\ding{51} & & &74.09 \\
       \ding{51}   & & &\ding{51} & &75.41  \\
       \ding{51}   &\ding{51} &\ding{51} & & &75.31  \\
       \ding{51}   &\ding{51} & &\ding{51} & &75.94 \\
       \ding{51}   &\ding{51} &\ding{51} &\ding{51} & &76.15 \\ 
       \ding{51}   &\ding{51} &\ding{51} & &\ding{51} &76.16  \\
       \ding{51}   &\ding{51} &\ding{51} &\ding{51} &\ding{51} &76.26  \\
    \bottomrule
    \end{tabular}}
    \label{tab:ablation}
\end{table}

\noindent
\textbf{Comparison of Feature Alignment Implementations.} 
Tab. \ref{tab:feature-align} compares various feature alignment implementations. 
On the premise of implementing feature alignment, it is discovered that all feature-based methods can define additional logit-based objectives on output to achieve positive results. This demonstrates that feature alignment is possible within a logit-based scheme and that the two are not mutually exclusive. 
Notably, the OFD connector is also a single-layer 1×1 convolution, which is consistent with our connector design. This connection method is lightweight and only considers the linear connection on the channel to align the dimension of the feature, allowing for high scalability. 
Based on the consistent connector structure, the comparison between Block and OFD can exactly reflect the increase of unified feature alignments (Fig. \ref{fig2}c) over direct alignment (Fig. \ref{fig2}b), indicating the source of our scheme's benefits.
Compared to other feature alignment schemes, our scheme obtains the best results, demonstrating that the proposed implicit feature alignment via logits is meaningful, and it allows for further investigation into the potential of feature alignment. \\

\noindent 
\textbf{Ablation Study of Training Objectives.} 
Experiments involving the ablation of training objectives are conducted, with the ablation components being added sequentially to determine their effects. 
The results are summarized in Tab. \ref{tab:ablation}. 
Vallina KD is applied to the logit distance $d^L$ for $L^S_{distill}$, $L^N_{distill}$, and $L_{cross}$ in this experiment.
The original student objectives utilized in the logit-based distillation methods, $L^S_{task}$ and $L^S_{distill}$, are defined based on the classification targets and the teacher's soft labels (Eq. \ref{eq:kd}). The objectives of the stepping stones, denoted $L^N_{task}$, $L^N_{distill}$, and $L_{cross}$, are defined by classification targets, teacher logits, and preceding step logits.
It can be found that $L^N_{distill}$ has a significant performance improvement when used in conjunction with other objectives, indicating that our implementation of logit-based feature alignment is practical and effective.
In addition, introducing $L^N_{task}$ can further enhance performance based on $L^N_{distill}$, demonstrating that the definition of task training objectives for stepping-stone can aid in the general functioning of the distillation framework.
Furthermore, the expected improvement in overall performance can be realized by incorporating the logit distillation between phases, $L_{cross}$. 
All objectives have contributed positively to the distillation framework and are compatible with one another.

\begin{figure}[tb]
  \centering
  \includegraphics[width=1.0\linewidth]{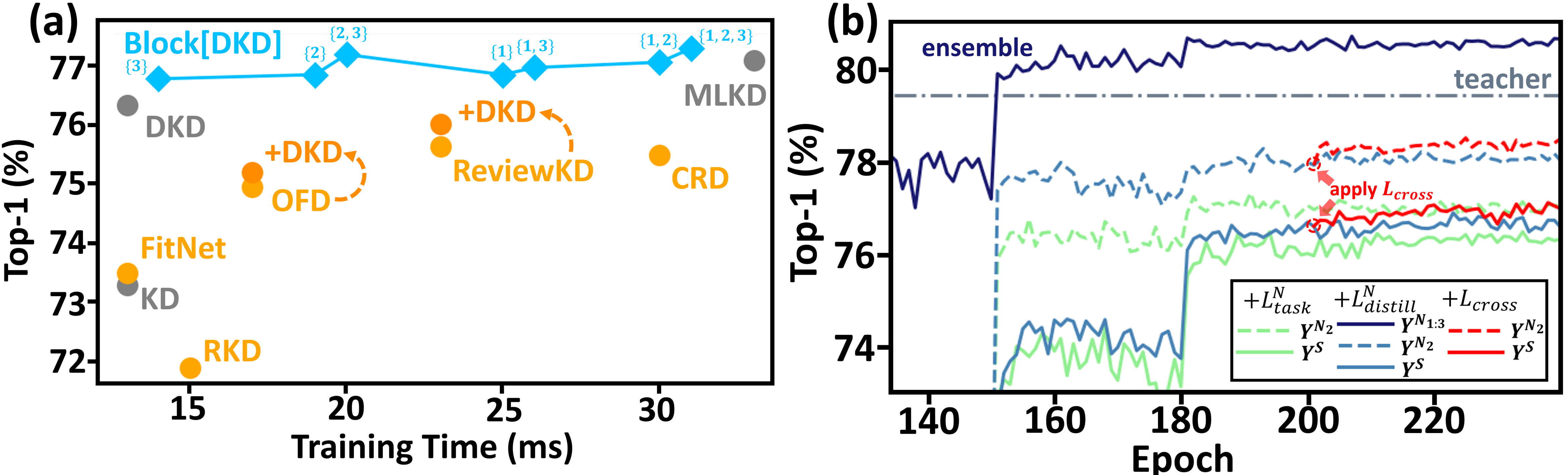}
   \caption{Visualization Results with the R32x4 \& R8x4 Pair on CIFAR-100. (a) Cost comparison. (b) Validation results during training. The green line represents results obtained by adding $L^N_{task}$ to the baseline; the blue line further adds $L^N_{distill}$; and the red line incorporates $L_{cross}$ at the 200$^{th}$ epoch upon the blue line.}
   \label{fig:fig_r}
\end{figure}

\noindent
\textbf{Discussion on Loss Combinations} 
The purpose of $L^N_{task}$ is to guide the stepping models towards task convergence, thereby facilitating collaborative work of $S$'s blocks and connectors with $T$'s blocks. Building upon this, $L^N_{distill}$ enhances stepping models and achieves feature alignment in the latent logits space. $L_{cross}$ leverages ensemble learning to derive superior soft labels, elevating the overall performance.
The visualized results align with our expectations (Fig. \ref{fig:fig_r}b). Adding $L^N_{distill}$ enhances branch training, improving $S$ and $N_2$ (green to blue). Incorporating $L_{cross}$ at the 200$^{th}$ epoch further extends the performance boundary (blue to red).

\noindent
\textbf{Training Overhead.} Tab. \ref{tab:train_cost} contrasts the training costs of various methods. 
All feature-based methods must contain additional parameters to coordinate the teacher and the student; since our framework only requires the extension of connectors, the number of parameters is relatively modest.
Due to the necessity of projecting the alignment features into the logit space, the training speed is slowed. Block-KD has similar training costs to CRD (Fig. \ref{fig:fig_r}a). However, a simple method can be employed to reduce the computational complexity by decreasing the required number of stepping stones (see Block[DKD]$^\dag$ in Tab. \ref{tab1} and Tab. \ref{tab2}), which may be considered to optimize deployment; the discussion is included in the following section.

\begin{table}[tb]
    \centering
    
    \caption{\textbf{Comparison of training costs under different schemes.} ResNet32×4 and ResNet8×4 are paired on CIFAR-100. Three indicators of concern in the training phase are listed: auxiliary parameters (MB), training duration (milliseconds per batch), and top-1 accuracy (\%).}
    \resizebox{0.45\textwidth}{!}{
    \begin{tabular}{c|ccc}
    \toprule
        & Time (ms)
        & Params (MB)
        & Top-1 (\%)
        \\   
        \midrule

        KD \cite{hinton_distilling_2015} &13	&0	&73.33  \\
        
        FitNet \cite{romero_fitnets_2015} &13	&0.065	&73.50  \\
        \midrule
        RKD \cite{park_relational_2019} &15	&0            &71.90  \\
        CRD \cite{tian_contrastive_2022} &30	&49.06	    &75.48  \\
        OFD \cite{heo_comprehensive_2019} &17	&0.332	    &74.95  \\
        ReviewKD \cite{chen_distilling_2021} &23	&6.87	    &75.63  \\
        DKD \cite{zhao_decoupled_2022} &13	&0	    &76.32  \\

        \midrule

        Block[DKD]$^\dag$ &20	&0.315	&\textbf{77.18}  \\
        Block[DKD]$  $ &31	&0.332	&\textbf{77.28}  \\
        \bottomrule
    \end{tabular}}
    \vspace{-10pt}
    \label{tab:train_cost}
\end{table}

\begin{table}[tb]
    \centering
    
    \caption{\textbf{Comparison of training costs under different settings of stepping-stones.} ResNet32×4 and ResNet8×4 are paired on CIFAR-100. The base refers to DKD \cite{zhao_decoupled_2022}, where stepping-stones $N_1$, $N_2$, and $N_3$ are added gradually for performance enhancement. Three indicators of concern in the training phase are listed: auxiliary parameters (MB), training duration (milliseconds per batch), and top-1 accuracy (\%).}
    \resizebox{0.45\textwidth}{!}{
    \begin{tabular}{c|ccc}
    \toprule
        & Time (ms)
        & Params (MB)
        & Top-1 (\%)
        \\   
        \midrule

        base &13	&0	&76.32  \\
        \midrule
        w/ $N_1$ &25	&\textbf{0.0161}	&\textbf{76.84}  \\
        w/ $N_2$ &19	&0.0635	            &\textbf{76.84}  \\
        w/ $N_3$ &\textbf{14}	&0.2520	    &76.77  \\

        \midrule

        w/ $N_1$,$N_2$ &30	&\textbf{0.0796}	&77.05  \\
        w/ $N_1$,$N_3$ &26	&0.2681	&76.96  \\
        w/ $N_2$,$N_3$ &\textbf{20}	&0.3154	&\textbf{77.18} \\
  
        \midrule
        w/ $N_1$,$N_2$,$N_3$ &31	&0.3315	& 77.28  \\
        \bottomrule
    \end{tabular}}
    \vspace{-10pt}
    \label{tab:train_cost_ab}
\end{table}

\noindent
\textbf{Pruning Stepping-stones}.
The implicit computation of stepping-stones increases training costs since features must be projected into the logits space through the teacher's back-end feature extraction blocks. The core goal of knowledge distillation is to produce a lightweight model so that training costs can be accepted to some extent \cite{wang_knowledge_2022}. Despite the fact that this issue is not critical, we consider mitigating it to illustrate the optimization field and application value of block-wise logit distillation. Note that all the initial stepping-stones are utilized to ensure the one-to-one correspondence between blocks; however, this restriction may be excessively stringent.
Therefore, as an optimization strategy, it makes sense to consider reducing the quantity of stepping stones. The outcomes of our experiments to prune stepping-stones on the resnet32×4/resnet32×8 pair are shown in Tab. \ref{tab:train_cost_ab}.
For the pair composed of resnet32×4 and resnet8×4, it is divided into three blocks during distillation; therefore, three stepping-stones are used to determine logits at this time, which are recorded as $N_1$, $N_2$, and $N_3$. 
Pruning on stepping-stones reveals that $N_1$ has a major impact on computing time, but that only a small number of parameters need to be introduced. This is primarily because the number of shallow channels is small, and thus the input and output dimensions of the connector are relatively smaller. However, the calculation of $N_1$ must traverse nearly all of $T$'s blocks, nearly doubling the calculation time. 
$N_3$ has the largest effect on the number of implemented parameters, but its additional calculation time is negligible, as expected. The calculation of $N_3$ is equivalent to leasing $T$'s classification to $S$, thus the calculation cost is very low; however, due to the large number of channels, the number of introduced parameters is the highest. 
$N_3$ performs marginally worse when only a single stepping-stone is used, while $N_1$ and $N_2$ perform similarly, and $N_2$ is recommended due to its faster processing speed.
In the case of two stepping stones, the combination of $N_2$ and $N_3$ achieves the highest performance and the fastest calculation speed. 
In conclusion, the middle stepping stone is recommended first when selecting stepping stones; shallow stepping stones (with smaller $i$) can be appropriately discarded when training time is a factor, and deep stepping stones (with larger $i$) can be appropriately discarded when the number of auxiliary parameters is a factor.

\noindent
\textbf{Why does $L_{cross}$ improve the Block-KD?}
$L_{cross}$ is a further optimization of logit-based feature alignment, which is simply extended within the framework of block-wise logit distillation in practice. The benefits it provides can be viewed from two perspectives. At first, the concept of ensemble learning may be introduced, in which stepping-stone models are regarded as separate students. Therefore, the knowledge acquired by all students can effectively guide each student to enhance their learning outcomes. This concept is known as student collaboration. Second, using the output of a previous stepping-stone as the target for a subsequent stepping model is consistent with the intuitive notion of feature alignment, where the focus of alignment has shifted away from the front-end student block and teacher block (that is, alignment between blocks $\{T_1,..., T_i\}$ and $\{S_1,..., S_i\}$, when $T$ and $N_i$ are paired). Instead, when the same student block is positioned at the front, there is a desire to align both the student's blocks and teacher's blocks in the middle (say, alignment between blocks $\{T_{i+1}, …, T_k\}$ and $\{S_{i+1}, …, S_k\}$, when $N_i$ and $N_k$ are paired), which represents an enhanced version of the original feature alignment.

\noindent
{\textbf{Potential Biases and Ethical Implications.} 
In the Block-KD framework, while knowledge distillation from a teacher model to a student model generally enhances performance, it is worth noting some subtle nuances. There could be a slight potential for the student model to inherit biases present in the teacher model, simply because it learns not only the explicit task at hand but also deeper patterns that may not be immediately evident. Additionally, the implicit nature of knowledge transfer through logit and feature alignment might make it a bit challenging to fully trace decision-making processes within the student model, which could be relevant in scenarios where understanding the basis of decisions is important. As a precautionary measure, while not dominant, ensuring that the distilled model performs fairly across various groups does merit attention. It’s also sensible to consider the source of the distilled knowledge; the teacher model’s quality can subtly influence the ethical standing of the student model. If the teacher model has not been thoroughly evaluated for potential ethical concerns, there's a slight chance that some of these might carry over to the student model.
}

\section{Conclusion}
\label{sec:conclusion}

This paper shows the potential of integrating logit and features. Focusing on two mainstream types in knowledge distillation, feature- and logit-based methods, we reveal that the essential distinction between the two is their explicit prior alignment to "dark knowledge". 
This work provides a novel perspective on feature alignment to interpret the underlying differences between logit and feature distillation, and extends the standard implementations to achieve feature alignment in logits space.
Furthermore, this work proposes a new distillation framework called block-wise logit distillation (Block-KD), which implements the alignment of feature-level "dark knowledge" based on the unified logit-based distillation with intermediate step logits generated by implicit stepping-stones.
We hope this study will inspire future research to explore KD from a broader, more comprehensive vantage perspective, utilizing both logits and features. 

\section*{acknowledgement}
This work was supported by the National Natural Science Foundation of China (Grant No. 62304203), the Natural Science Foundation of Zhejiang Province, China (Grant No. LQ22F010011), and the ZJU-UIUC Center for Heterogeneously Integrated Brain-Inspired Computing.

\bibliographystyle{plainnat} 
\bibliography{IEEEabrv,reference}

\end{document}